# Identification of morphological fingerprint in perinatal brains using quasi-conformal mapping and contrastive learning

**Boyang Wang 1, Weihao Zheng 1,2 , Ying Wang 1, Zhe Zhang 3,4 , Yuchen Sheng 5,\* and Minmin Wang 2,6,\***

1. Gansu Provincial Key Laboratory of Wearable Computing, School of Information Science and Engineering, Lanzhou University, Lanzhou, China.
2. College of Biomedical Engineering and Instrument Science, Zhejiang University, Hangzhou, China.
3. Institute of Brain Science, Hangzhou Normal University, Hangzhou, China.
4. School of Physics, Hangzhou Normal University, Hangzhou, China.
5. School of Foreign Languages, Lanzhou Jiaotong University, China
6. Binjiang Institute of Zhejiang University, Hangzhou, China.
* Correspondence: shengyc_lzjtu@163.com (Y.S.); 12015005@zju.edu.cn (M.W.)

**Abstract:**

The morphological fingerprint in the brain is capable of identifying the uniqueness of an individual. However, whether such individual patterns are present in perinatal brains, and which morphological attributes or cortical regions better characterize the individual differences of ne-onates remain unclear. In this study, we proposed a deep learning framework that projected three-dimensional spherical meshes of three morphological features (i.e., cortical thickness, mean curvature, and sulcal depth) onto two-dimensional planes through quasi-conformal mapping, and employed the ResNet18 and contrastive learning for individual identification. We used the cross-sectional structural MRI data of 682 infants, incorporating with data augmentation, to train the model and fine-tuned the parameters based on 60 infants who had longitudinal scans. The model was validated on 30 longitudinal scanned infant data, and remarkable Top1 and Top5 accuracies of 71.37% and 84.10% were achieved, respectively. The sensorimotor and visual cortices were recognized as the most contributive regions in individual identification. Moreover, the folding morphology demonstrated greater discriminative capability than the cortical thickness, which could serve as the morphological fingerprint in perinatal brains. These findings provided evidence for the emergence of morphological fingerprints in the brain at the beginning of the third trimester, which may hold promising implications for understanding the formation of in-dividual uniqueness in the brain during early development.

## 1. Introduction

Significant progress has been achieved in understanding brain development and organization due to advances in neuroimaging and neuroscience (Giedd et al., 2010 [1]; Mills et al. 2014 [2]). There is growing awareness of the importance of differences in brain structure and function among individuals. These inter-individual variations may underlie differences in cognitive abilities, emotional processing, and susceptibility to neurological and psychiatric disorders (Dannlowski et al., 2012 [3]; Karama et al., 2014 [4]; Schmaal et al., 2017 [5]). Some researchers have explored the potential of brain function and structure for the usage of biometric identification (Bassi et al., 2018 [6]; Chen et al., 2018 [7]). While prior investigations have revealed that the adult brain exhibited stable structural and functional fingerprints for representing individual differences (Finn et al., 2015 [8]; Menon et al., 2019 [9]), it remains unclear whether the fingerprint in the brain emerges as early as during the third trimester—a critical period marked by the explosive growth of cortical anatomy and rapid establishment of structural and functional connectome.

Magnetic Resonance Imaging (MRI) is a non-invasive imaging technique known for its capacity to yield high-resolution images for the examination of both brain structure and function. This technology has been widely applied to study infant brains, such as the early developmental patterns of morphology, microstructure, fiber tracts, and brain connectomes (Liu et al., 2021 [10]; Liu et al., 2021 [11]; Zheng et al., 2023 [12] ; Zheng et al., 2023 [13]). Studies have characterized individual differences in human brains through anatomical and functional connectivity. For example, the white matter tractography in adult brains could serve as an effective fingerprint for the identification of individuals (Yeh et al., 2016 [14]), and the identification accuracy of functional connectivity increases with age from childhood to adulthood (Vanderwal et al., 2021 [15]). Furthermore, recent studies have attempted to investigate the fingerprint in perinatal brains and achieved recognition rates of 62.22% and 78% based on structural (Ciarrusta et al., 2022 [16]) and functional (Hu et al., 2022 [17]) connectivity, respectively. These results demonstrated the fact that individual uniqueness of brain connectome emerges during early brain development. Similar as brain connectivity, cortical morphology could also serve as a valuable fingerprint for individual recognition (Wachinger et al., 2015 [18]; Aloui et al., 2018 [19]), which achieved remarkable performance that superior to the functional connectivity in differentiating adult individuals (Tian et al, 2021 [20]). In recent years, some studies have begun to explore cortical folding patterns for infant subject identification. For instance, Duan et al. successfully identified 1- and 2-year-old infants by using cortical folding information (i.e., curvature, convexity, and sulcus depth) of the corresponding neonate (Duan et al.,2019 [21]; Duan et al.,2020 [22]). Nevertheless, it remains unclear whether the individual variations in human brain morphology already appear as early as the beginning of the third trimester.

Deep learning methods can achieve more advanced feature representations in brain images. Multiple deep learning models have been applied to resolve challenges in brain image analysis in recent years. For example, the convolutional neural network (CNN) has been utilized for the detection of brain lesion (Chen et al., 2020 [23]) and white matter abnormalities (McKinley et al., 2019 [24]), as well as for the diagnosis of brain disorders (Esmaeilzadeh et al., 2018 [25]; Qureshi et al., 2019 [26]). Convolutional networks based on the whole-brain cortical surface (Mostapha et al. 2018 [27]) have offered new perspectives for studying human brain MRI and cortical morphology. However, 3D-CNN methods are often unsuitable for small-sample datasets due to the large number of parameters and high computational demands. On the other hand, 2D-CNN methods have lower computational requirements and can be more easily embedded into a mature model architecture. T. W. Meng et al. have introduced Teichmüller Extremal Mapping of POint clouds (TEMPO), a quasi-conformal mapping method for conformally mapping a simply-connected open triangle mesh to a 2D rectangle space [28]. The TEMPO method can effectively preserve conformality, reducing the loss of local features and geometric structures in the mapping of the original point cloud. Leveraging deep learning methods enables the automatic acquisition of more high-dimensional features of cortical morphology through multilevel nonlinear transformations, thereby simplifying the model's principles and procedures via end-to-end learning. Therefore, we utilized quasi-conformal mapping to project the 3D brain mesh onto a 2D plane and employed 2D-CNN to extract individual cortical morphological feature representations.

The present study aims to validate the existence of morphological fingerprints in perinatal brains. Each hemispheric surface of an individual subject was inflated to a sphere and was then projected to a 2D plane through TEMPO. We propose a contrastive learning framework based on the pretrained ResNet18 encoder to recognize an individual at term-equivalent age by using his or her brain MRI acquired at birth. The attention mechanism is incorporated to fuse features from different partitions generated from the 3D-to-2D mapping. The effectiveness of each morphological feature is assessed to demonstrate their contributions to the individual fingerprint in perinatal brains.

## 2. Materials and Methods

### 2.1. Dataset

#### 2.1.1. Dataset Description

Imaging data used in this study was obtained from the Developing Human Connectome Project (dHCP-release 3, https://biomedia.github.io/dHCP-release-notes/acquire.html), which was generously supported by the European Research Council (ERC). Ethical approval for the project was granted by the UK Health Research Authority (Research Ethics Committee reference number: 14/LO/1169), and written parental consent was diligently obtained for both the imaging procedures and data release.

All participants received MRI scans at the Evelina Newborn Imaging Centre, located at St Thomas' Hospital in London, UK. The scans included the acquisition of structural, diffusion, and functional data, which was performed using a 3 Tesla Philips Achieva system (running modified R3.2.2 software) [29] . T2-weighted (T2w) multi-slice fast spin-echo images were obtained in both sagittal and axial slice stacks. These images had an in-plane resolution of 0.8x0.8mm² and 1.6mm slices, with an overlap of 0.8mm. The imaging parameters were as follows: repetition time (TR) = 12000ms, echo time (TE) = 156ms, SENSE factor of 2.11 for axial and 2.60 for sagittal slices. Additionally, a 3D MPRAGE scan was performed with 0.8mm isotropic resolution.

The dataset included 783 infants in total, comprising 682 infants who underwent MRI scanning one time and 101 infants who were scanned twice (scanned at approximately 1-2 weeks after birth and term-equivalent age, respectively) or more. We selected the T2 images of 772 infants (90 of them have longitudinal scans), and images of 11 participants were manually excluded due to poor quality. Detailed demographic information of the selected infants is given in Table 1.

**Table 1.** Demographic information of the infants included in this study.

| Groups | Median birth age (weeks) | Birth age range (weeks) | Median scan age (weeks) | Scan age range (weeks) | Male/ Female |
|---|---|---|---|---|---|
| Cross-sectional cohort | 39.86 | [23.0-43.57] | 41.14 | [26.86-45.15] | 370/ 312 |
| Longitudinal cohort (1st scan) | 31.29 | [23.57-40.14] | 34.22 | [26.71-42.71] | 48/42 |
| Longitudinal cohort (2nd scan) | 31.29 | [23.57-40.14] | 41.29 | [35.57-44.86] | 48/42 |

### 2.1.2. Dataset Processing

Image preprocessing followed the pipeline proposed by the dHCP [30]. In summary, the process first involved bias correction and brain extraction of the motion-corrected T2-weighted image. This was followed by segmenting the brain into different tissue types using Draw-EM algorithm. White-matter mesh was then extracted and expanded to fit the pial surface. The cortical thickness was estimated based on the Euclidean distance between the white and pial surface. We generated the inflated surfaces from the white surface, which were then projected to a sphere for surface registration. The mean surface curvature and sulcal depth (mean surface convexity/concavity) were respectively estimated from the white surface and from inflation.

### 2.2. Teichmüller Mapping

Despite the promising spherical convolutional neural network (SCNN) models on (Cohen et al., 2018 [31]; C. Esteves et al., 2018 [32]; C. Jiang et al., 2019 [33]), the SCNNs often suffer from substantial computational complexity, e.g., extensive parameters, which make them difficult to converge on small sample datasets. Therefore, we projected the 3-dimensional spherical mesh of the brain surface to 2-dimensional and employed the convolutional neural network (CNN) model for analysis.

To achieve this goal, a quasi-conformal mapping method (Teichmüller Mapping) was employed, which is a geometric transformation method based on complex variational functions. Conformal mapping facilitates the mapping of one space to another while preserving local angles between intersecting curves or surfaces, thereby ensuring the preservation of shapes and angles within a small region of the original space. In general, its formula representation is as follows:

$$f * ds_N^2 = \lambda ds_M^2$$

where M and N are two Riemann surfaces, f is the mapping function: M→N, and λ is a positive scalar function.

A generalization of conformal mapping is quasi-conformal mapping, which allows for a certain degree of angle and shape distortion to adapt to more complex data structures and transformations. Teichmüller Mapping is such a type of quasi-conformal mapping that can induce uniform conformality distortions in the target point cloud, thereby preserving stable relative positions and local shapes. The general formula for a Teichmüller mapping (T-map) can be expressed as follows: let M and N be two Riemann surfaces, and $f: M \rightarrow N$ be a quasi-conformal mapping. If it is associated with a quadratic differential $q = \phi dz^2$, where $\phi: M \rightarrow \mathbb{C}$ is a holomorphic function, and its associated Beltrami coefficient is of the form $\mu(f) = k\frac{\phi}{|\phi|}$, where k<1 is a constant, and the quadratic differential q is non-zero and satisfies $||q||_1 = \int_{S^1} |\phi| < \infty$, then f is called a Teichmüller mapping (T-map) associated with the quadratic differential q.

In the present study, we used the TEMPO method for projection. Specifically, we adopt an approach of segmenting the hemispherical plane along the x=0 plane in the coordinate system and subsequently applying the TEMPO method separately to the two hemispheres for mapping. Then, we interpolated the mapped mesh to obtain a 2-dimensional matrix. Although this approach sacrificed the continuity of the data around the segmentation curve, it significantly reduced the area distortion from mapping deformation and preserved the overall continuity of other areas, offering a more desirable outcome for our study. The images projected to the 2D plane are shown in the first column of Figure 1.

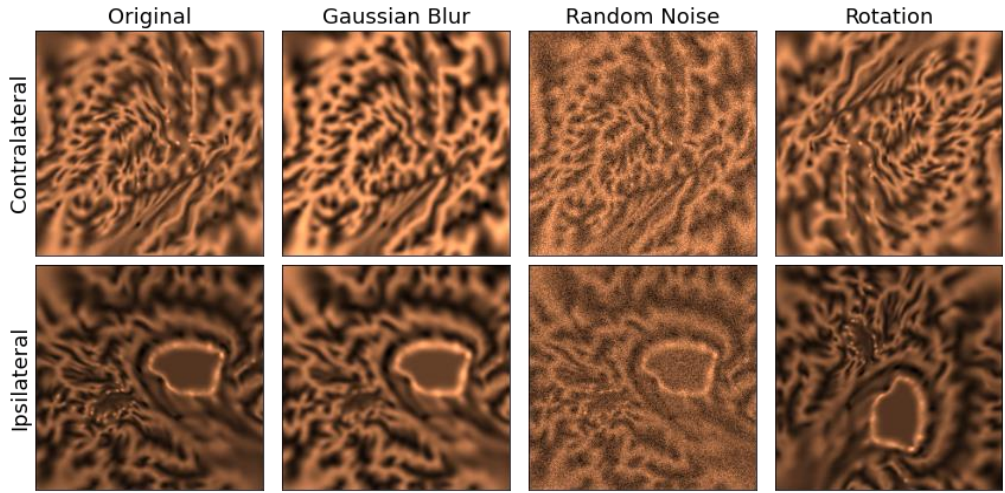

**Figure 1.** Three data augmentation methods used on the dataset.

## 2.3. Data Augmentation

Data augmentation is a commonly employed technique to address the issue of insufficient data and lack of data diversity. Its primary objective is to expand the dataset by applying diverse transformations and processing to the original data. In the context of images, various augmentation methods have been established, including random cropping, rotation, flipping, random noise, Gaussian blur, and color transformations such as brightness, contrast, and saturation adjustments (Cubuk et al., 2018 [34]; Shorten et al., 2019 [35]; T. Chen et al., 2020 [36]). We used data augmentation to expand single sampling data into sample pairs, which can be used for model pre-training. We employed rotation, random noise, and Gaussian blur on both the one-shot and two-shot data, as shown in Figure 1.

## 2.4. Model Architecture

As illustrated in Figure 2-(a), we present a contrastive learning-based feature extraction framework. We aligned the 3D grids of the left and right brain hemispheres onto spheres, and then partitioned each sphere into two hemispheres. Subsequently, we employed the TEMPO method to map the spherical grids onto planar grids and compute four 3×224×224 matrices through interpolation. These cortical morphological matrices were then fed into the feature extraction module to extract cortical morphological feature fingerprints of individual subjects (Figure 2-(b)). These fingerprints (vectors of length 512) were ultimately utilized for calculating pairwise similarities for individual identification. Notably, in the feature extraction process, we employed a channel-wise attention mechanism-based excitation module for fusing features from different brain partitions and weight allocation to feature map channels, as shown in Figure 2-(c).

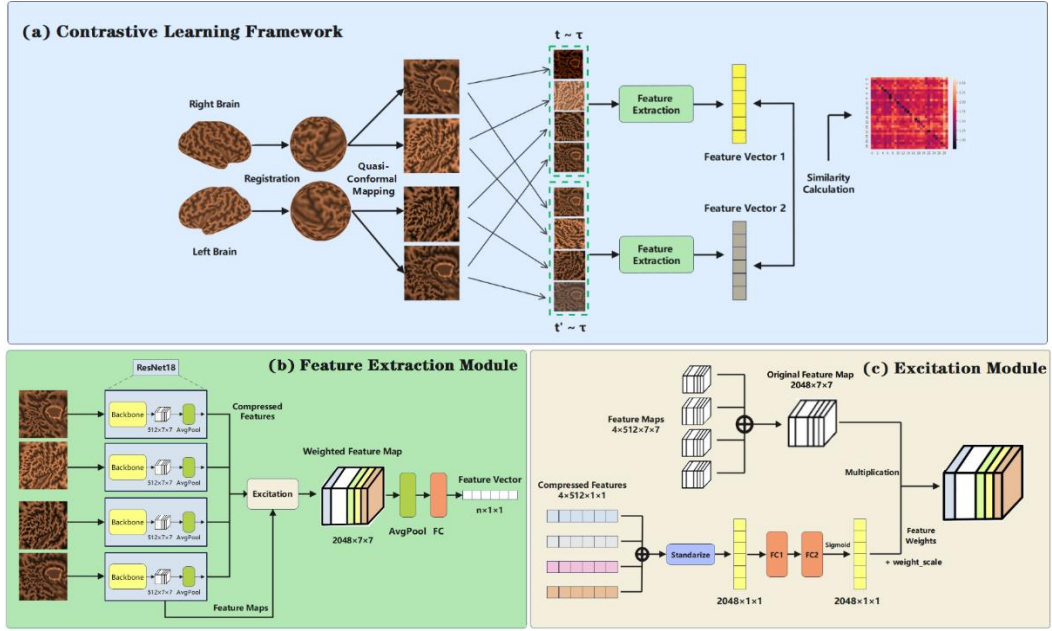

**Figure 2.** A simple contrastive learning framework for perinatal cortical morphology fingerprint. (a) demonstrates the process of utilizing original data for data augmentation and applying it within a contrastive learning framework; (b) shows the workflow within the green feature extraction module in (a), which extract feature representation from brain maps; (c) shows the workflow within the grey excitation module in (b) which calculates and utilizes the channel weights for channel attention-based feature learning and fusion.

### 2.5. Contrastive Learning

Contrastive learning is a self-supervised method for training models to learn meaningful data representations. It works by comparing the similarity between pairs of data samples. In essence, it creates pairs of positive and negative samples and uses a loss function to make the feature representations of positive pairs more similar while making those of negative pairs more dissimilar. A notable framework for contrastive learning is SimCLR (short for Simple Contrastive Learning of Representations) [34] , which used the NT-Xent loss function to maximize the similarity of positive pairs and minimize the similarity of negative pairs in the feature space. The NT-Xent loss function is defined as:

$$L = -\frac{1}{N}\sum_{i=1}^{N}\left(\log\left(\frac{\exp(-\|z_i - z_i^+\|/\tau)}{\sum_{j=1, j\neq i}^{N}\exp\left(-\|z_i - z_j^+\|/\tau\right)}\right)\right)$$

where N is the batch size, $\|z_i - z_i^+\|$ is the Euclidean distance between the i-th sample and its positive sample, $\|z_i - z_j^+\|$ is the Euclidean distance between the i-th sample and the j-th sample's corresponding positive sample. τ denotes a temperature parameter.

In this study, we used the SimCLR as the main contrastive learning framework to extract morphological feature representations of individual differences in the perinatal cerebral cortex. We constructed positive sample pairs by using multiple data augmentation methods on single sampling data and used the other samples in the same batch to construct negative pairs. Specifically, we adopted a contrastive loss similar to the classical Siamese network [37]. We constructed multiple negative sample pairs while adopting a non-softmax computation of the loss function. The cross-entropy loss that was separate for the positive and negative sample pairs was calculated and summed up. The loss function was defined as follows:

$$L = \sum_{i=1}^{N} \left( (1 - y_i) * \text{dist}_i^2 + y * \text{clamp}_{\min}(m - \text{dist}_i, 0)^2 \right)$$

where

$$\text{dist}_i = \| x_{1i} - x_{2i} \|, \text{clamp}_{\min}(a, b) = \begin{cases} a, & \text{if } a \geq b \\ b, & \text{if } a < b \end{cases}$$

and N was the batch size, $y_i$ was the label indicating whether the i-th sample pair was a positive or negative sample, $x_{1i}$ and $x_{2i}$ represented the two samples of the i-th sample pair respectively, and m was an artificially set margin.

In this framework, we used the ResNet18 backbone (He et al., 2016 [32]) as the brain map encoder, as shown in Figure 2-(b). For each morphological feature, four sets of projection maps (medial and lateral projections of the left and right brain, respectively) per sample were fed into the parameter-sharing ResNet18 to extract four feature maps.

### 2.6. Excitation Module

The excitation module aimed to enhance the overall representation capability of the feature vectors to the morphological features of the samples through the channel attention mechanism, as shown in Figure 2-(c). It achieved the fusion of features from four different partitions. Similar to the SE-Nets (Hu et al., 2018 [38]), the channel weights were learned by two fully connected layers that mapped the feature vector to a channel attention vector. This channel attention vector was applied to the input feature map to weigh the channels. However, different from the previous literature, our excitation module optimized the assignment of the weight to three channels of morphological features and four channels (partitions) of projection maps simultaneously, which was arranged before the final layer of the neckbone network. Adding a hyperparameter weight_scale was intended to restore the weights to a distribution with a mean of 1 or approximately 1 to maintain the consistency of the data scale. The feature vector of the sample was obtained by pooling and a simplified projection through a fully connected layer. In addition to the excitation method, we employed other two fusion techniques: a decision-level fusion approach based on a voting mechanism and a feature-level fusion method using a two-layer Multilayer Perceptron (MLP). The former averaged similarity matrices calculated on four partitions and then made identification decisions based on the averaged similarity matrix. The latter utilized a two-layer MLP to map the input features. Specifically, we flattened a 4×2048 feature vector and then passed it through the MLP, resulting in a feature vector of length 2048.

### 2.7. Training Procedure and Evaluation Metrics

For the training process, we initially excluded the excitation module and employed the original structure during the training of the ResNet18 backbone. The network parameters were trained by using primarily augmented sample pairs for pre-training and several two-shot data for fine-tuning, as illustrated in the stage1 of Figure 3. After that, we froze the ResNet18 and updated the parameters in the excitation module and the FC layer (stage2 of Figure 3). Finally, the trained feature extraction module was applied to the twice scanned samples. We calculated and compared the similarity of the feature vectors between two time points of the same subject (self-self similarity) and the feature similarity between a neonate at birth and other infants scanned at term-equivalent age (self-other similarity).

The Euclidean distance was used to measure the similarity between individual brains (Eickhoff et al.,2005 [39]; Al-Saffar et al., 2020 [40]). We utilized the Top 1 accuracy (where self-similarity surpasses all self-other similarities) and Top 5 accuracy (where self-similarity ranks among the top five similarities between itself and all other samples) as metrics to assess the effectiveness of the model.

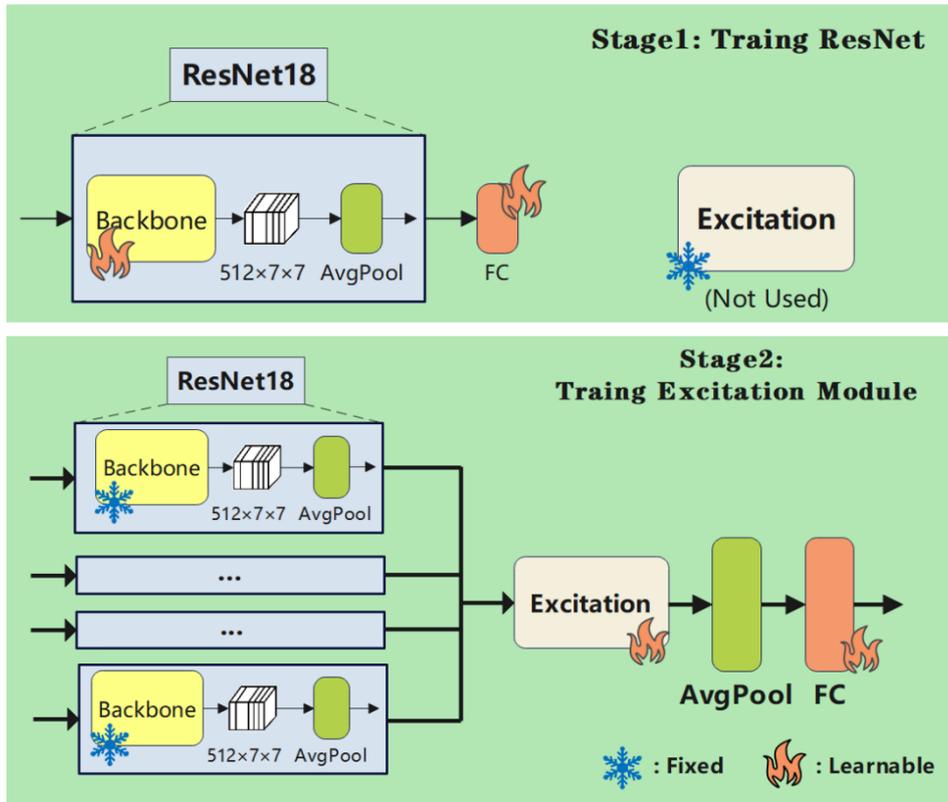

Figure 3. A two-stage learning strategy for model parameters

## 3. Results

### 3.1. Experimental Setup

We used the ResNet18 backbone network and its pre-trained network parameters provided by Pytorch on the ImageNet dataset for initialization. To train the ResNet18 network, we augmented all single-sampled data to create positive and negative sample pairs. These pairs were then used for the initial training of ResNet18. Subsequently, we performed fine-tuning of ResNet18 and trained the feature fusion module. The training process employed a learning rate of 5e-4 with a momentum of 0.9 and a weight decay of 5e-5. Each training session comprised 8 epochs and the SGD optimizer was used. For all experiments, we conducted triple-fold cross-validation using twice sampled data (i.e., the real sample pairs). In each fold, we used 60 and 30 real sample pairs for training and testing, respectively. We repeated the experiments for 30 rounds to ensure robustness and obtained weighted accuracies for identification.

### 3.2. Experiment Results

#### 3.2.1. Individual Recognition

Our model achieved a notable Top 1 accuracy of 71.37% and a Top 5 accuracy of 84.10%. These results suggested that the self-similarity of the prenatal brains between birth and term-equivalent ages was higher than all self-other similarities or was among the top 5 highest. We also compared the accuracies derived from different backbone models (i.e., ResNet18, 34, and 50) and fusion strategies for fusing features from left lateral, left medial, right lateral and right medial brain. (e.g., excitation, voting and MLP). The excitation method outperformed other fusion methods. The voting strategy that integrated judgments from different partitions also demonstrated high Top 1 accuracy of 71.37% and Top 5 accuracy of 83.90%. The method using a MLP for feature fusion achieved the Top 1 accuracy of 68.70% and Top 5 accuracy of 79.63%. Overall, the excitation method exhibited superior performance. In addition, more network layers did not give better performance, as shown in Table 2.

**Table 2.** Recognition accuracy with different encoders and training epochs

| Backbone | Number of total epochs | Excitation method Top1 accuracy (%) | Decision-level method Top1 accuracy (%) |
|---|---|---|---|
| ResNet18 | 16 | 71.37 | 71.37 |
|  | 32 | 70.20 | 69.21 |
|  | 64 | 68.05 | 68.16 |
| ResNet34 | 16 | 68.43 | 68.56 |
|  | 32 | 65.43 | 66.10 |
|  | 64 | 60.50 | 60.10 |
| ResNet50 | 16 | 62.67 | 60.78 |
|  | 32 | 63.44 | 60.22 |
|  | 64 | 60.56 | 59.77 |

### 3.2.2. Contributions of Morphological Features and Brain Regions

To explore the contributions of each morphological feature to the recognition task, we conducted single-channel comparison experiments for the three morphological features: curvature, thickness, and sulcus. The obtained results are presented in Figure 4. The curvature achieved the highest Top 1 and Top 5 accuracies of 79.13% and 86.27%, respectively, surpassing the combination of all the three channels. The sulcus feature achieved a Top 1 accuracy of 67.10% and Top 5 accuracy of 78.40%. However, cortical thickness did not exhibit any discriminative power, which not only achieved the accuracies lower than 50% but may also reduce the accuracies up to 25% when combined with other features.

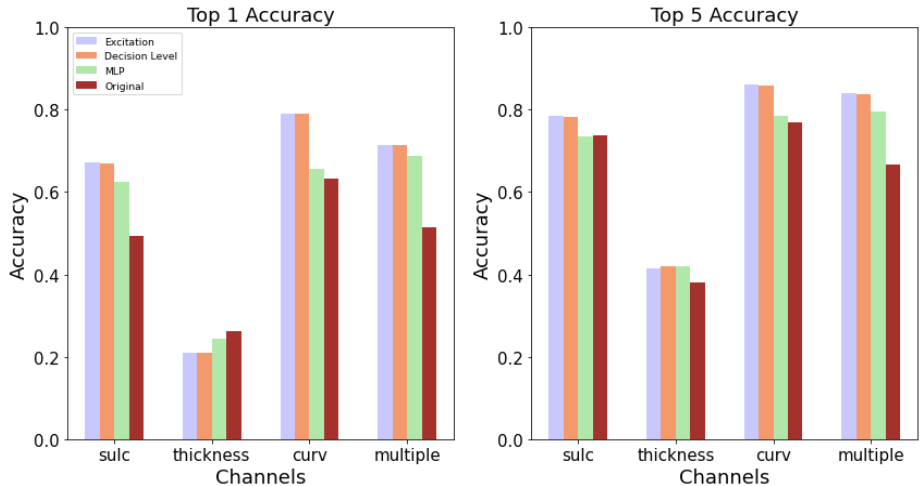

**Figure 4.** Top 1 and Top 5 accuracy comparison between different fusion methods and different feature channels selected.

We then explored the contributions of different brain regions in the recognition task. We utilized the weights derived from the excitation module as the contribution weights of brain regions shown in Figure 5-A, and the weights of brain regions were mapped back to the cortical surface (as illustrated in Figure 5-B). The parietal and occipital cortices of the left hemisphere and the antero-medial and postero-lateral temporal cortices of the right hemisphere showed higher attentional weights relative to other regions, suggesting these regions may have evident morphological differences among individuals.

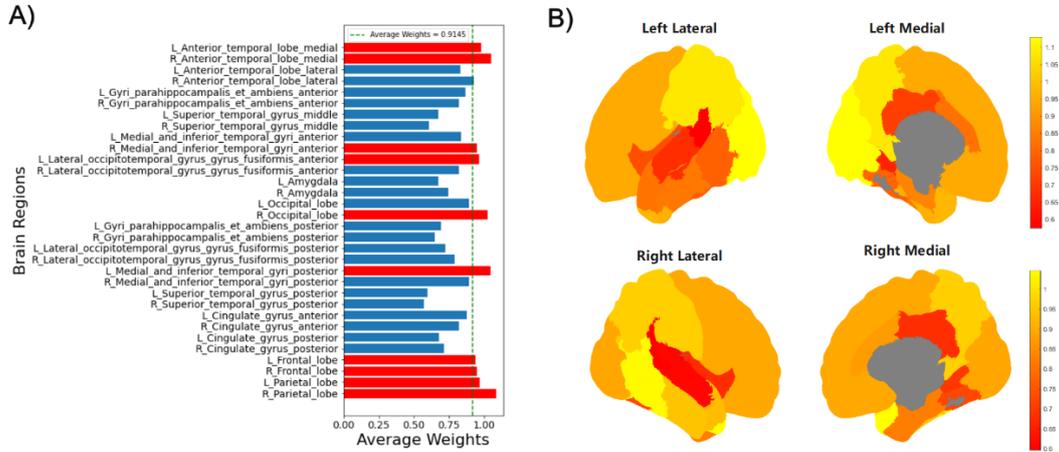

**Figure 5.** The average brain map weights of each brain region when using all three cortical morphological features. The vertical axis in A) represents different brain region labels, where L and R represent left and right. The red bars in A) are the brain regions with the top ten highest weights, and the green dashed line indicates the average weight of the whole brain. The average weights of each brain region in 2D brain atlases are shown in B). The gray area represents the corpus callosum, and attention weights are not calculated for it.

### 3.2.3. Ablation Studies

To further validate the effectiveness of our model, we conducted ablation experiments under the following conditions: (1) without using a contrastive learning framework to train encoders; (2) without using augmented data for pre-training; and (3) without using the excitation module to fuse multiple features. Without using the contrastive learning framework, the performance of a ResNet encoder pretrained only on ImageNet is quite poor, achieving only 10% Top1 accuracy. The training within the contrastive learning framework significantly enhanced the model's performance. The utilization of data augmentation and the excitation model also improved the model's performance, increasing Top1 accuracy by over 5% and over 50%, respectively. The details are presented in Table 3.

**Table 3.** Recognition accuracy under different experimental operation settings

| Operations | Top1 Accuracy(%) | Top5 Accuracy(%) |
| --- | --- | --- |
| A). Contrastive Learning + Pre-training + Excitation | 71.37 | 83.90 |
| B). Without contrastive learning | 10.00 | 20.00 |
| C). Without pre-training on augmented data | 66.10 | 80.23 |
| D). Without Excitation Model | 19.33 | 65.90 |

Note: A) followed the complete model pipeline. In B), a pre-trained ResNet18 with ImageNet weights was used as an encoder without further contrastive learning training. C) involved the training using only twice sampled data, without using augmented samples from single time sampling. D) omitted the excitation Module for fusing brain map features from four regions and directly performed concatenation instead.

## 4. Discussion

In this study, we examined the existence of morphological fingerprints in neonatal cerebral cortex through a deep learning model. We achieved a remarkable Top 1 accuracy of 71.37% in the individual recognition. This suggested that certain cortical morphological fingerprints are already formed as early as the beginning of the third trimester and maintained stable during the perinatal period, which can serve as an effective fingerprint for recognizing individual neonate.

Training the ResNet18 encoder with a contrastive learning approach significantly improved the recognition accuracy, demonstrating the effectiveness of the framework in learning cortical morphological features. Additionally, the incorporation of the excitation module based on attention mechanisms also enhanced the Top 1 accuracy by over 50%, outperforming both the comparative voting and MLP methods. We speculated that the improvement might be attributed to the excitation module enabling the model to prioritize brain regions with significant individual variability, reducing the impact of other regions on individual recognition and thus enhancing identification accuracy (Bodapati et al., 2021 [41]). Moreover, the excitation module enabled the direct assessment of the contribution of different brain regions, providing a convenient way to analyze the recognition contribution rates of different brain areas. Pretraining the model by constructing additional sample pairs using data augmentation further boosted the Top 1 accuracy by over 5%, effectively improving the efficiency of using limited samples in cases of insufficient data. This improvement has been confirmed in several classification task based on brain images (Garcea et al. 2022 [42]). The presence of these distinctive features enables the model to extract individual-specific information for each subject, achieving high-accuracy individual recognition.

Our findings indicated that both cortical curvature and sulcal depth exhibited individual variations, and the former achieved the best recognition rate, emphasizing the crucial role of the cortical folding morphology in characterizing individual differences at early developmental stage. On the other hand, cortical thickness did not exhibit discriminative power to individual recognition. Previous studies have suggested that the individual variability of cortical folding patterns has been established at term age (Duan et al., 2019 [43]), while the cortical thickness showed relative longer maturation period. Specifically, the cortical thickness typically undergoes rapid growth shortly after birth, peaking around 14 months after birth and subsequently decreases (Wang et al., 2019 [44]). Therefore, we hypothesized that the earlier maturation of folding patterns imparts greater individual variability to cortical morphology, leading to higher accuracy in individual recognition. Additionally, sensitivity to noise may be another potential factor leading to lower recognition efficiency of thickness features. Furthermore, we observed that the primary cortices associated with somatosensory and visual functions carried higher attention weights, indicating greater inter-individual differences in these regions. Given that the primary regions experienced more pronounced development than high-order cortex in the second trimester (Gilmore et al., 2018 [45]; Duan et al., 2019 [43]), we speculated that the high attention weights assigned to the primary cortex may be attributed to their stable morphology maturity throughout the third trimester.

There were several limitations for this study. First, the longitudinal dataset was relatively small, which only contained 90 infants. The generalizability of this model and the reproducibility of our findings should be examined on a larger independent dataset. Second, although the conformal transformation could ensure point cloud continuity by using the deformation errors, it may also lead to area distortion during projection, e.g., the compression of the area farther away from the segmentation curve and the expansion the area closer to the segmentation curve. Such distortion may potentially influence the recognition accuracy. Some novel methods for addressing these challenges should be developed in the future, although it is out of the scope of our work.

## 5. Conclusions

Our study showed that cortical folding morphology, especially the curvature, significantly contributed to individual variations of perinatal brains rather than cortical thickness. Moreover, regions with high individual differences mainly concentrated in the primary cortex. These findings offered the first evidence of the existence of individual morphological fingerprints in neonatal brains as early as the beginning of the third trimester, which maintained relatively stable during the perinatal period.

## Funding

This work was funded by the National Natural Science Foundation of China (grant number 62202212), "Pioneer" and "Leading Goose" R&D Program of Zhejiang (grant number 2023C03081) and the Fundamental Research Funds for the Central Universities (grant number 226-2023-00091).